\documentclass{article}

\usepackage{PRIMEarxiv}
\usepackage{xcolor}   
 
\usepackage[utf8]{inputenc} 
\usepackage[T1]{fontenc}    
\usepackage{hyperref}       
\usepackage{url}            
\usepackage{booktabs}       
\usepackage{amsfonts}       
\usepackage{nicefrac}       
\usepackage{microtype}      
\usepackage{orcidlink}
\usepackage{lipsum}
\usepackage{fancyhdr}       
\usepackage{graphicx}       
\graphicspath{{media/}}     
\usepackage{amsmath}
\usepackage{multirow}
\usepackage{graphicx}

\usepackage{colortbl}
\title{Self-Attention Networks Enable Compact Latent Space Modelling of Dynamic Optical Fibre Matrices}

\author{Yijie~Zheng\orcidlink{0000-0002-6513-1584}\textsuperscript{1},Robert~J.~Kilpatrick\textsuperscript{2},Qing Yang\textsuperscript{3,4}, David~B.~Phillips\textsuperscript{2},George Gordon\orcidlink{0000-0002-7333-5106}\textsuperscript{1}\\
\textsuperscript{1}Optics and Photonics research group, University of Nottingham, UK\\
\textsuperscript{2}University of Exeter, UK\\
\textsuperscript{3}State Key Laboratory of Extreme Photonics and Instrumentation, College of Optical Science and Engineering\\ International Research Center for Advanced Photonics, Zhejiang University, Hangzhou, China\\
\textsuperscript{4}Research Center for Humanoid Sensing, Zhejiang Lab, Hangzhou, China\\
\texttt{george.gordon@nottingham.ac.uk}}

\begin{document}
\maketitle

\begin{abstract}
Multimode optical fibres, hair-thin strands of glass that guide light, promise next-generation ultra-thin medical endoscopes for sub-cellular imaging deep within the body. However, their thinness means that during light transmission there is inherent scrambling of images. Conventional methods correct this scrambling using a pre-calibrated linear matrix to capture the fibre's effect on light but dynamic changes caused by movement, temperature variations, and measurement non-linearities when the fibre tip is inaccessible limit the effectiveness in many real-world applications. Nonlinear models, such as neural networks, can better capture dynamics but conventional convolutional architectures fall short because they assume local pixel correlations -- an assumption invalidated by the long-range correlations in fibre matrices arising from both physical modal coupling and the flattening of multidimensional optical fields into matrix representations. To overcome this, we introduce a new approach using self-attention layers to dynamically transform varying fibre matrices into compact, low-dimensional latent-space representations, well-suited for later downstream tasks such as image reconstruction or bending estimation. Our method significantly enhances sparsity of fibre matrix models, with participation ratios between 0.01 and 0.11, and allows for reconstructing the original matrices with less than 10\% error. The learned transformation preserves invertibility and remains compatible across different modal bases. Additionally, we demonstrate that our model tolerates dynamic transmission matrix perturbations and is compatible with experimentally measured data.
\end{abstract}

\section{INTRODUCTION}
Lensless endoscopic imaging using ultra-thin multi-mode fibres (MMFs) offers a state-of-the-art method for minimally invasive \textit{in vivo} imaging, allowing access to deep body regions such as the brain or blood vessels for advanced biomedical imaging \cite{psaltis2016imaging}. MMFs efficiently transmit light along their length due to a tailored refractive index profile, which is modelled physically by linear differential equations (Maxwell's equations). When spatially discretised, these equations become linear matrix equations, with input and output fields represented as complex-valued vectors and the effect of the fibres represented by complex-valued \textit{transmission matrices (TMs)}:
\begin{equation}
    \mathbf{E_{out}} = \mathbf{T} \mathbf{E_{in}}
    \label{eq:1}
\end{equation}
\noindent where $\mathbf{E_{in}} \in \mathbb{C}^{N}$ is a discretized and column-wise stacked vector representing the input field at one end of the fibre, $\mathbf{E_{out}} \in \mathbb{C}^{N}$ is the equivalent discretized output field, and $\mathbf{T} \in \mathbb{C}^{N\times N}$ is the complex-valued TM of the fibre. The scalar $N$ denotes the number of spatial modes the fibre supports, akin to the number of independent pixels in transmitted images. The field representation basis is flexible, accommodating canonical (pixel), Fourier, or Hadamard bases for example, which in practice, changes in measurement basis can obscure structure and complicate generalisable modelling.

Conventionally, linear matrix equations suffice to reconstruct static fibre TMs from input-output pairs measured using wavefront modulation \cite{cheng2022long, kuschmierz2018self,stasio2015light,weiss2018two,woo2021dynamic,woo2022optimal,zhao2021parameter}, speckle-correlated intensity reconstruction\cite{caravaca2021optical,stasio2016calibration,fan2021high,porat2016widefield} and compressive sensing\cite{caravaca2019hybrid,caravaca2023single,amitonova2018compressive}. These methods are considered static and linear since they infer a TM based on input-output pairs of a static fibre. However, in practical applications like medical imaging, fibres experience constant movement and temperature variations that dynamically alter the TM. Furthermore, obtaining input-output image pairs for calibration is impractical, as it requires bulky optical components at the fibre tip, which is deep within the body.

Thus, effective real-time fibre characterisation methods must model a dynamically changing TM, relying solely on single-end measurements. These measurements are typically nonlinear, involving forward TM estimation ($\mathbf{T}$) from quadratic-form round-trip measurements ($\mathbf{T}^\top\mathbf{T}$) \cite{gordon2019characterizing,wen2023single}. Previous approaches leverage feedback from guide stars \cite{li2017focus,li2021memory,weiss2018two}, tracked beacon sources \cite{farahi2013dynamic,wen2023single}, or reflective structures \cite{gu2015design,chen2020remote,gordon2019characterizing} affixed to the fibre tip. To complement this, there are also several physical models designed to model dynamic TMs under physical perturbations and wavelength modulation \cite{matthes2021learning,lee2023efficient}.  However, these models become intractable when modelling multiple types of perturbations that must be estimated using a limited set of measurements. There is therefore a need to develop a more flexible model for dynamic TMs that can be better fitted to measured data.

Trainable neural networks offer a promising alternative for modelling complex fibre behaviours. Convolutional neural networks (CNN) have been widely applied to image transmission and wavefront shaping through mostly static optical fibres \cite{rahmani2018multimode,caramazza2019transmission,fan2019deep,fan2021learning,borhani2018learning,resisi2021image}. CNNs are well-suited to static fibre image reconstruction scenarios, which are essentially image reconstruction problems, because they assume a strong inductive bias specific to images, namely that neighbouring pixels are highly correlated. To better capture the characteristics of MMFs, CNNs can be modified to encode phase information, using complex-weighted networks for improved speckle representation \cite{moran2018deep} or incorporating image transformers to enhance reconstruction quality \cite{wu2023high}. 

\begin{figure}[t]
    \centering
    \includegraphics[width=1\linewidth]{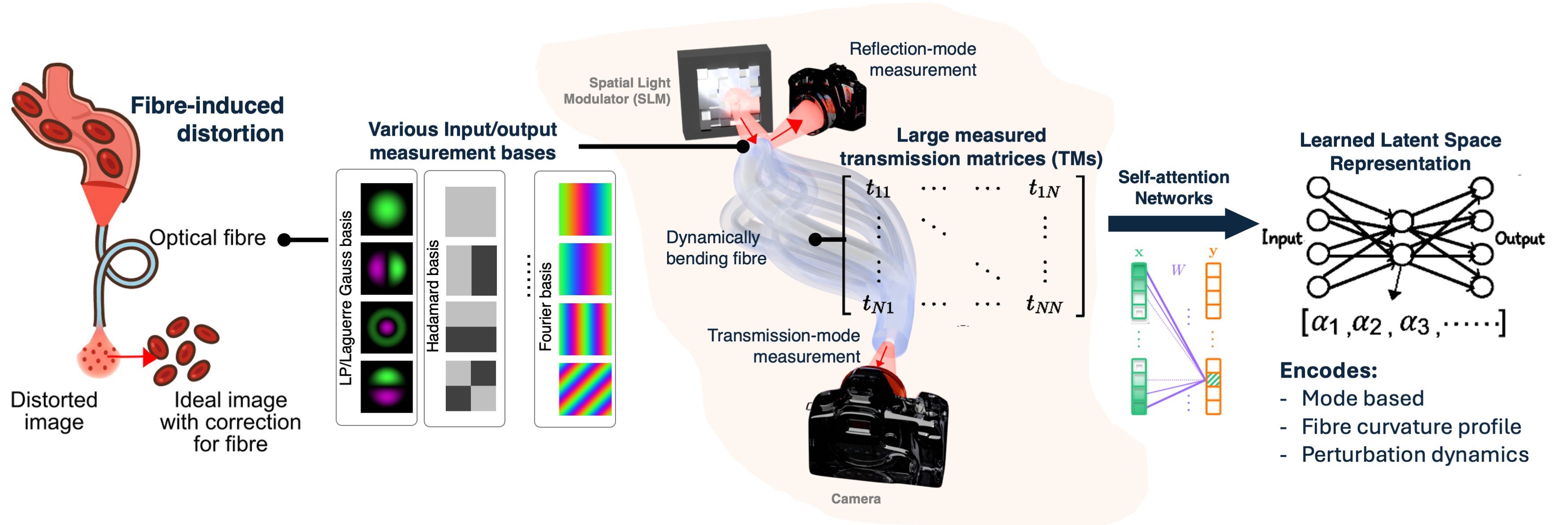}
    \caption{Concept of Self-Attention for Modelling Optical Fibres. Left to right: 1. Biomedical Imaging Motivation: Optical fibre imaging has the potential to enable new biomedical applications in ultra-narrow, hard-to-reach areas of the body. However, dynamic correction for fibre-induced distortion is essential for achieving clear images. 2. Challenges in Fibre Modelling: varying input/output measurement bases, combining reflection and transmission measurements, modelling fibre dynamics under bending or perturbation, capturing large transmission matrices (TMs) dealing with high dimensionality. 3. Self-attention Networks Modelling Approaches: Conventional convolutional neural networks (CNNs) are limited in this context, where they capture only short-range correlations. In contrast, self-attention networks, which underpin modern Transformer architectures used in most Large Language Models, can capture long-range correlations and complex fibre dynamics, making them well-suited to this task. Here, long-range correlations include both non-local modal coupling and representation-induced non-locality caused by flattening two-dimensional input/output field structures into matrix indices.} We validate the self-attention model by demonstrating its ability to compress large fibre TMs into a compact latent space via an autoencoder. This learned latent representation encodes mode bases, fibre curvature, and perturbation dynamics, enabling high-fidelity TM reconstruction and laying the foundation for downstream tasks such as image manipulation, missing data imputation, and real-time distortion correction.
    \label{fig:summary}
\end{figure}

For dynamically varying fibres, a more expressive model is needed to physically approximate the changing TM itself rather than focussing on image reconstruction. Dynamic TM modelling with CNNs has demonstrated effectiveness in learning prior distributions over TM perturbations, leading to improved generalization when modelling dynamically varying TMs \cite{zheng2023single}. However, these models often require prohibitively large numbers of parameters and lack robustness to changes in basis representation, such as misalignments within the optical system. Consequently, neural networks for dynamic TM modelling should meet particular requirements: 
\begin{enumerate} 
\item Be memory efficient to manage TMs supporting typical MMF image resolutions, achieved by transforming matrices to a sparse basis that can be reduced to a low-dimensional latent space.
\item Support complex numbers.
\item Handle arbitrary different measurement bases (i.e., eigenbases), offering a degree of basis invariance or equivariance. 
\item Be sufficiently expressive to capture and model dynamic behaviour.
\end{enumerate}

Therefore, a neural network model for dynamically varying TMs should convert a range of perturbed TMs across different basis representations into a compact latent space, following with the `compression is learning' principle \cite{deletang2023language}, as shown in Figure \ref{fig:summary}. In this way, compression-based neural network models, such as autoencoders, are employed to map large TMs into a learned compact latent space vector representation. This latent representation encodes critical information about the fibre’s input mode bases, curvature profile, and dynamic perturbations, enabling more efficient and accurate modelling of dynamically bending fibres.



In this paper, we investigate whether a self-attention-based neural network, when directly applied to TMs, effectively models dynamic TMs by reducing them to a compact latent space, producing sparse representations suitable for reconstruction and downstream inference tasks. Self-attention, which captures long-range pairwise dependencies between elements in an input sequence (usually represented as a matrix), is widely used in natural language processing tasks such as machine translation \cite{tang2018self,shaw2018self}, text summarization \cite{liu2019text,xu2020self}, and sentiment analysis \cite{medhat2014sentiment}, where understanding contextual relationships is essential for accurate predictions. Recently, self-attention mechanisms have also been applied to physical systems, such as in generative diffusion models for image reconstruction from on-sensor measurements on nanophotonic array cameras \cite{chakravarthula2023thin}. In this context, self-attention functions as a learnable, input-dependent (i.e., nonlinear) basis transformation, converting input matrices into output matrices, which learns input-dependent interactions between modes, which is consistent with the non-local coupling structure of fibre TMs. Specifically, 'long-range correlations' refer to non-local dependencies between input and output modal channels in the TM representation, rather than only to spatial correlations in the measured optical field. Such correlations can arise from modal crosstalk, mode-group coupling, bend-induced mixing, and perturbation-dependent redistribution of amplitude and phase across modes. Importantly, modes that are physically coupled are not necessarily adjacent in the matrix index ordering or in the chosen measurement basis. This non-locality is a consequence of how TMs are represented. Along one matrix dimension, the ordering of input modes is partly arbitrary: even when modes have natural radial and azimuthal indices, this two-dimensional structure must be mapped onto a one-dimensional matrix index. Along the other dimension, a two-dimensional output field or image is commonly flattened into a one-dimensional vector, so neighbouring pixels or field components in the physical image plane may become separated in the matrix representation. Moreover, unitary changes of input or output basis can mix rows and columns and therefore change the apparent locality of correlations without changing the underlying fibre physics. A CNN is effective for local neighbourhood features, but its convolutional kernels impose a local inductive bias. In contrast, self-attention allows each TM column or modal token to interact directly with all other tokens, making it better suited to learning non-local modal dependencies in the transformed representation. This capability therefore enables compact latent space representations, supporting efficient dynamic modelling and reconstruction.

We validate our approach by enabling compact latent space representations across diverse simulated and experimentally-measured TM datasets, including randomly generated forward and round-trip TMs, physically modeled TMs under perturbations, and experimental TM datasets. First, we show that the performance of CNNs degrades following basis transformations, especially when the transformed matrices are dense, and therefore do not meet our first criterion. Next, we demonstrate that Fully Connected Neural Networks (FCNNs) achieve better performance following parameter transformation, although their parameter count scales unfavourably with TM size violating the second criterion. Finally, introducing self-attention mechanisms markedly improves upon both CNNs and FCNNs by enabling sparse matrix representations across all datasets, with enhanced scalability.

We believe that this approach to TM modelling and compression could facilitate the deployment of single-ended fibre imaging, paving the way for hair-thin imaging devices that fit within a needle \cite{gordon2019characterizing,zheng2023single,li2021compressively}. More broadly, our method could apply to the efficient modelling of matrix data in other scattering media contexts \cite{rotter2017light,chaigne2014controlling} and perturbation conditions such as wavelength shifts \cite{lee2023efficient}.
\section{METHODS}
\subsection{Diverse TM datasets}
We generate four diverse datasets (with different sparsity properties) of $78 \times 78$ complex-valued perturbed TMs: randomly-generated forward TMs, randomly-generated round-trip TMs and physically modeled TMs. For the randomly-generated forward TMs, $\mathbf{T_f}\in\mathbb{C}^{78\times 78}$, we construct the simulation model \cite{zheng2023single} to recreate typical properties (e.g. high sparsity, high power intensity along the main diagonal and sub-diagonals, condition number 3-10) observed in real fibre TMs. These are inherently sparse in their original basis.  For the randomly generated round-trip TMs, $\mathbf{T_r}\in\mathbb{C}^{78\times 78}$, based on a nonlinear reflection-mode fibre imaging system \cite{gordon2019characterizing,zheng2023single}, we use the forward TM, $\mathbf{T_f}$, to compute a round-trip TM as:
\begin{equation}
    \mathbf{T_r}=\mathbf{T_f}^T\mathbf{R}\mathbf{T_f}
\end{equation}
\noindent where $\mathbf{R}\in\mathbb{C}^{78\times 78}$ is the round-trip matrix that numerically represents reflection off the distal fibre facet, which contains a highly scattering reflector. We use a highly scattering/random reflector in order to generate dense round-trip TMs with strong mode mixing and favourable conditioning for single-ended recovery. A flat mirror is not excluded in principle: previous theory shows that flat and random reflectors can leave a comparable phase ambiguity (a unitary matrix) which may be resolved using multi-wavelength probing \cite{gordon2019characterizing} or single-wavelength if the fibre TM is perfectly unitary \cite{gu2015design}. However, random or structured reflectors are generally observed to provide more stable inversion because they introduce richer mixing and more distinct reflector eigenvalues \cite{gordon2019characterizing}. As a result, these round-trip TMs are inherently dense in their original basis. 

For each simulated dataset, we generate 22,000 TMs and split them into training, validation and test sets with an 8:2:1 ratio. The validation set is expected to provide unbiased evaluations and stopping criteria and the test set aims to examine the generalization performance of the model on unseen data.

For the physically modelled TM dataset, we generated TMs using the segmented fibre model, in which each perturbed fibre is represented as a cascade of short fibre sections with specified length and bend radius. Specifically, the simulated fibre parameters were fixed to a core radius of \(8~\mu\mathrm{m}\), numerical aperture \(\mathrm{NA}=0.22\), and Poisson ratio \(0.17\). Simulations were performed at a wavelength of \(532~\mathrm{nm}\). Fibre lengths were sampled over \(0.6\text{-}1.2~\mathrm{m}\), covering the metre-scale MMF lengths relevant to minimally invasive endoscopic imaging configurations. Bend radii were sampled over \(20\text{--}100~\mathrm{mm}\), corresponding to curvatures of approximately \(10\text{--}50~\mathrm{m}^{-1}\). This range was chosen to include both weakly and more strongly perturbed fibre states while retaining physically plausible multimode-fibre configurations \cite{ploschner2015seeing}. This means the ensembled TM, $\mathbf{T_e}\in\mathbb{C}^{78\times 78}$, can be therefore calculated by the product of the individual segmented TMs, $\mathbf{T_{s_1..z}}\in\mathbb{C}^{78\times 78}$ of the fibre:
\begin{equation}
    \mathbf{T_e}=\mathbf{T_{s_1}}\mathbf{T_{s_2}}...\mathbf{T_{s_z}}
\end{equation}

Finally, the experimental TMs, used in this chapter are collected from the experimental setup provided in \cite{wen2023single}, which involves modulating the input light and recovering the output. Specifically, for input modulation, collimated light is directed into a multi-mode fibre through an objective lens and a 4$f$ optical configuration after reflecting off a digital micro-mirror device (DMD). In this setup, the input fields are modulated using the Hadamard basis, which provides an orthogonal set of inputs. Next, to recover the complex output, images of the MMF’s output facet are firstly captured with a movable calibration module. The light then exiting the MMF is combined with a reference signal via a beam splitter, where the reference light is delivered to the camera through a single-mode fibre. Using off-axis holography, the combined signals are recorded by a CCD camera after magnification by a microscope objective. This optical experimental setup therefore enables precise measurement of the TM, providing valuable insights into the propagation characteristics of light through the MMF. Using this approach, experimental TMs are measured for 164 different bending conformations of the fibre. From these measurements, 78 rows and 78 columns are selected to create matrices of suitable size for training the model. 

\subsection{Basis transformation model}
Our objective is to compare several different approaches to basis transformation and thus present a versatile framework for modelling the fibre TMs in efficient coordinate systems as judged by their sparsity of representation. We construct and train the models using Tensorflow 2.1 running on an NVIDIA GeForce RTX 3090 GPU. The Adam optimizer was used with a learning rate of 0.001 in a decay rate of 1e$^{-5}$.
\begin{figure}[!htbp]
    \centering
    \includegraphics[width=0.8\linewidth]{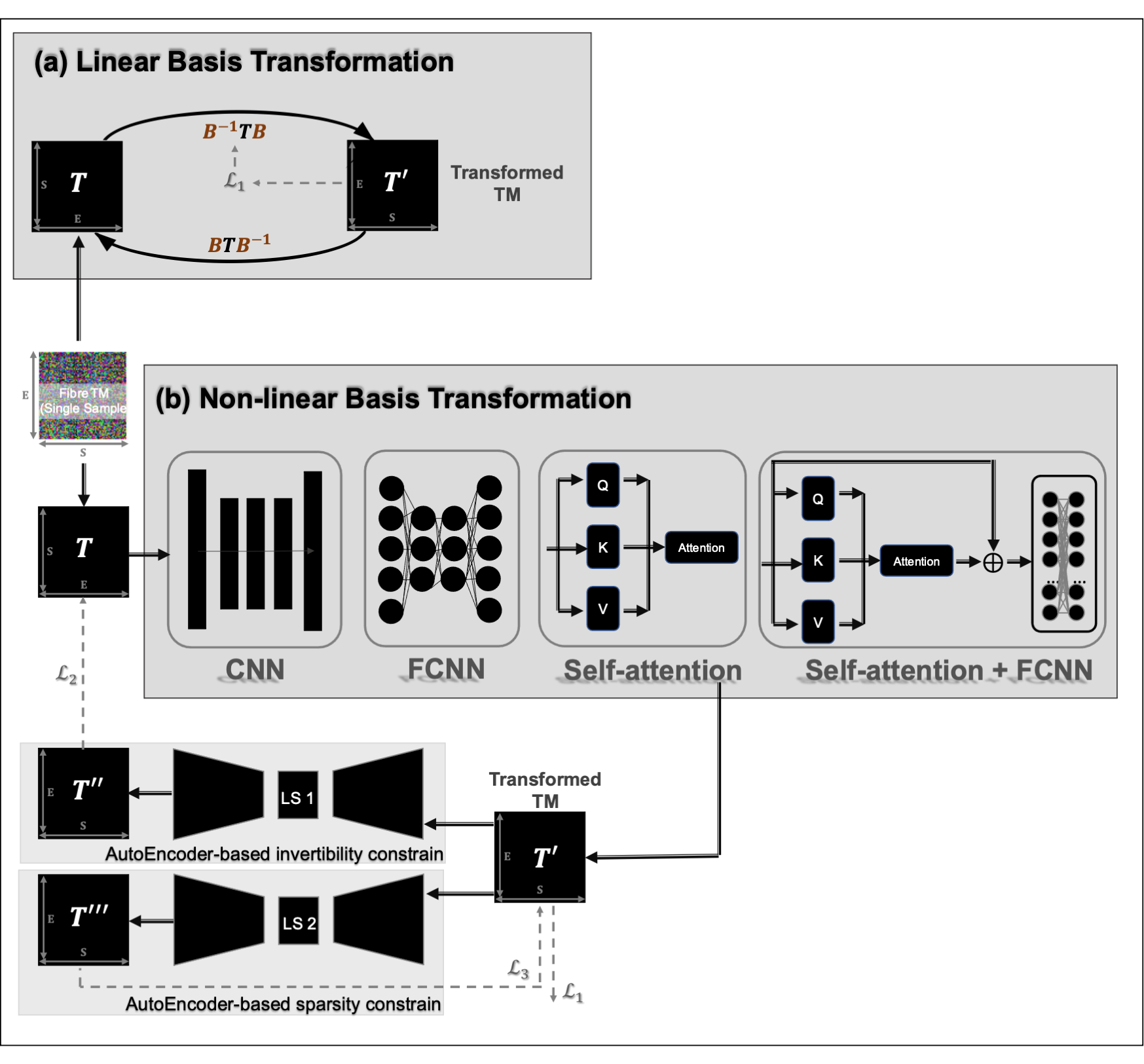}
    \caption{Basis transformation models (a) Linear similarity transformation-based model, with a lossless invertible process defined. (b) Non-linear transformation using different neural network architectures, including CNN, FCNN, self-attention only and self-attention-based FCNN. Two types of autoencoder-based constraints are defined for the purpose of sparsity representation and model invertibility.}
    \label{fig:methods}
\end{figure}
\subsubsection{Linear similarity transformation-based model}
We first construct a linear similarity transformation-based model as shown in Figure\ref{fig:methods}(a), to linearly convert TMs into a new coordinate system based on the theory of similarity transformation \cite{hahn2003similarity}. Specifically, we developed a differentiable custom layer, where $\textbf{B}\in\mathbb{C}^{78\times 78}$ is a non-singular square basis matrix that transfers sets of TMs (i.e.\ $\mathbf{T_{1...m}}\in\mathbb{C}^{78\times 78}$) to a new basis (i.e.\ $\mathbf{T'_{1...m}}\in\mathbb{C}^{78\times 78}$), while preserving the matrices properties (determinant, trace and eigenvalues):
\begin{equation}
    \mathbf{T'_{1...m}} = \mathbf{B}^{-1}\mathbf{T_{1...m}}\mathbf{B}
    \label{eqn1}
\end{equation}

This is a differentiable operation where the model is expected to learn the basis matrix \textbf{B} during the training, in order to enable a low-dimensional representation (i.e. increased sparsity) in the new basis,$\mathbf{T'_{1...m}}\in\mathbb{C}^{78\times 78}$, by minimizing an $\ell_1$ norm loss function ($\mathcal{L}_{linear}$):
\begin{equation}
    \mathcal{L}_{linear} = \sum_{i=1}^{m}\| \mathbf{T'_i} \|_1
\end{equation}
This linear similarity-based transformation model in turn, ensures a lossless invertible transformation of TMs to the original basis based on the basis matrix $\mathbf{B}$ learned by the model during the training.

\subsubsection{Non-linear transformation models}
We next construct non-linear basis transformation frameworks performed with four different non-linear neural networks models, namely CNN, FCNN, single self-attention layer and self-attention-based FCNN as shown in Figure\ref{fig:methods} (b). To feed the model, we convert the input sets of TMs, $\mathbf{T_{1...m}}\in\mathbb{C}^{78\times 78}$ into sets of real-valued TMs $\mathbf{T_{1...m}}\in\mathbb{R}^{156\times 156}$ by separating the real and imaginary parts\cite{hile1990matrix}. We expect to obtain corresponding sets of TMs, $\mathbf{T'_{1...m}}\in\mathbb{R}^{156\times 156}$, as the output that are low-dimensional representations (i.e.\ highly sparse). 

The first model is a CNN model that contains four Conv2D and MaxPooling layers with 190,965 trainable parameters and uses LeakyRelu activations with a negative slope value defined as 0.3.

The second model is an FCNN comprising two 1D dense layers with the LeakyRelu activation function (negative slope value defined as 0.3), with a comparable trainable parameters (203,534) as that of the CNN model.

The third and fourth models make use of self-attention mechanisms. Here, we encode the input information of each TM to three sets of interpretable matrices, Query ($\mathbf{Q_{1...j}}\in\mathbb{R}^{156\times 156}$),  Key ($\mathbf{K_{1...j}}\in\mathbb{R}^{156\times 156}$) and  Value ($\mathbf{K_{1...j}}\in\mathbb{R}^{156\times 156}$), with local interactions defined within each column. The output of the attention mechanism can be therefore updated values ($\mathbf{T'_{1...j}}\in\mathbb{R}^{156\times 156}$) for each element within the column groups accordingly:
\begin{equation}
\mathbf{Attention_{1...j}}=\mathrm{softmax}\left[\frac{\mathbf{Q}_{1...j}\mathbf{K}^T_{1...j}}{\sqrt{d_k}}\right]\mathbf{V}_{1...j}
\end{equation}

The third model directly learns this self-attention transformation, with the softmax as the only nonlinearity. The final model (details shown in Section 3.1) adds two FCNN layers following the self-attention to enhance the modeling capability.

\subsubsection{Autoencoder-based constraint}
By optimising these non-linear transformation models using only the $\ell_1$ norm, we may inadvertently lose inherent properties of the matrices. For example, the model might produce a transformation that maps all input values to zero.

To address this, we introduce two autoencoder-based constraints. The first autoencoder enforces invertibility of the learned transformation. It takes the transformed TM, $\mathbf{T'_{1...m}}\in\mathbb{R}^{156\times 156}$, as the input, maps it to a latent representation denoted LS1, and reconstructs the original TM (i.e. $\mathbf{T_{1...m}}\in\mathbb{R}^{156\times 156}$) as $\mathbf{T''_{1...m}}\in\mathbb{R}^{156\times 156}$ by minimizing the reconstruction loss, $\mathcal{L}_2$, during the training process. LS1 therefore measures whether the information required to recover the original TM is preserved after transformation.

We also enforce an additional sparsity constraint using a second Autoencoder model as shown in Figure \ref{fig:methods}, where the Encoder is designed to maximally compress the transformed matrices ($\mathbf{T'_{1...m}}\in\mathbb{R}^{156\times 156}$) into a low-dimensional latent space (LS2) via a `bottleneck' layer. Training this second Decoder ensures the LS2 contains the necessary information to reconstruct the transformed TM and is optimized by minimizing the reconstruction loss, $\mathcal{L}_3$. Therefore, the total loss of the whole framework can be defined as:
\begin{equation}
    \mathcal{L}_{\mathrm{total}} = \underbrace{\alpha\frac{1}{m}\sum_{i=1}^{m}\| \mathbf{T'_i} \|_1}_\mathrm{\mathcal{L}_1}+
    \underbrace{\frac{1}{m}\sum_{i=1}^m(\mathbf{T''_i}-\mathbf{T_i})^2}_\mathrm{\mathcal{L}_2}+
    \underbrace{\frac{1}{m}\sum_{i=1}^m(\mathbf{T'''_i}-\mathbf{T'_i})^2}_\mathrm{\mathcal{L}_3}
\end{equation}
\noindent where $\alpha=0.2$ is the coefficient for the direct sparsity constraint term. Both $\mathcal{L}_2$ and $\mathcal{L}_3$ use a custom loss-function that is invariant to global phase shifts of the matrices and also use a 2$\times$2 real number matrix representation to encode complex numbers \cite{zheng2023single}, thus meeting our second criterion of handling complex numbers.

\subsection{Evaluation metric}
To assess the effectiveness of transformation of the TMs, $\mathbf{T_{1...m}}\in\mathbb{C}^{78\times 78}$, we measure the mean participation ratio \cite{van1999localization} of the transformed TMs and define a metric, $p$, to quantify the ratio of occupied matrix elements (i.e. represent the sparsity), where $\mathbf{T_j}$ represents $j^{th}$ element of each TM: 

\begin{equation}
    p = \frac{(\sum_{i=1}^m\sum_{j=1}^{78\times78}|\mathbf{T_j}|^4)^{-1}}{78\times78}
    \label{metric1}
\end{equation}

Additionally, we construct an autoencoder model to downsample the transformed TM into a low-dimensional LS2 which is expected to contain the important features that can reconstruct the transformed TM. The compressibility (i.e.\ sparsity) of the original TMs, can be determined by:
\begin{equation}
    \mathrm{LS_{Ratio}} = \frac{1}{m} \sum_{i=1}^m\frac{{\texttt{Size of LS2}_i}}{156\times 156}
    \label{metric2}
\end{equation}

To further evaluate the invertibility of the transformed TMs, we construct a separate autoencoder model, where the decoder is expected to reconstruct the original TMs from LS1. We calculate mean squared error across the test dataset (mean error) between the reconstructed TMs and target TMs:
\begin{equation}
    \mathrm{mean error} = \frac{1}{m} \sum_{i=1}^m \sum (\mathbf{T''_i}-\mathbf{T_i})^2
    \label{metric3}
\end{equation}

\subsection{Cross-basis similarity of transformed TMs}

To assess whether the learned transformation preserves cross-basis structural consistency, we compare transformed TMs obtained from the same underlying fibre state expressed in different measurement bases. Specifically, a given TM $\textbf{T}$ is first represented in multiple measurement bases (LP/PIM, Fourier, and Hadamard) via unitary basis transformations. Each basis-specific matrix is then independently passed through the same trained self-attention-based FCNN network, resulting in a set of transformed matrices $\textbf{T}'^{(b)}$, where $b$ denotes the measurement basis.

To quantify similarity across bases, we compute the cosine similarity between vectorised transformed TMs,
\begin{equation}
\mathrm{sim}\!\left(\mathrm{vec}(\textbf{T}'^{(b_1)}),\,\mathrm{vec}(\textbf{T}'^{(b_2)})\right)
=
\frac{\mathrm{vec}(\textbf{T}'^{(b_1)}) \cdot \mathrm{vec}(\textbf{T}'^{(b_2)})}
{\|\mathrm{vec}(\textbf{T}'^{(b_1)})\|\,\|\mathrm{vec}(\textbf{T}'^{(b_2)})\|},
\end{equation}
where $\mathrm{vec}(\cdot)$ denotes flattening the matrix into a single vector. This metric measures the alignment of transformed representations independently of their absolute scale. High cosine similarity indicates that the learned transformation maps different basis representations of the same physical fibre state to nearby points in the transformed representation space.

As a control, cosine similarity is also computed between vectorised transformed TMs obtained from randomly paired transmission matrices. This ensures that any observed similarity across bases is not due to a degenerate mapping that produces nearly identical transformed representations for distinct transmission matrices. All similarity statistics are evaluated over the test dataset and reported as mean $\pm$ standard deviation.

This procedure isolates the effect of the learned transformation itself and provides a basis-agnostic quantitative measure of whether the transformed representation captures intrinsic structure of the fibre TM rather than overfitting to a particular measurement basis. The same transformation network and similarity metric are used consistently across all datasets and basis combinations to ensure the unbias comparison.

\section{RESULTS}
In this section, we investigate whether a learned transformation can reveal physically meaningful, sparse representations of fibre transmission matrices that are robust to different fibre mode bases and dynamic perturbations.
\subsection{Achieving sparsity using Self-attention + FCNN network}
\begin{figure}[!htbp]
    \centering
    \includegraphics[width=1\linewidth]{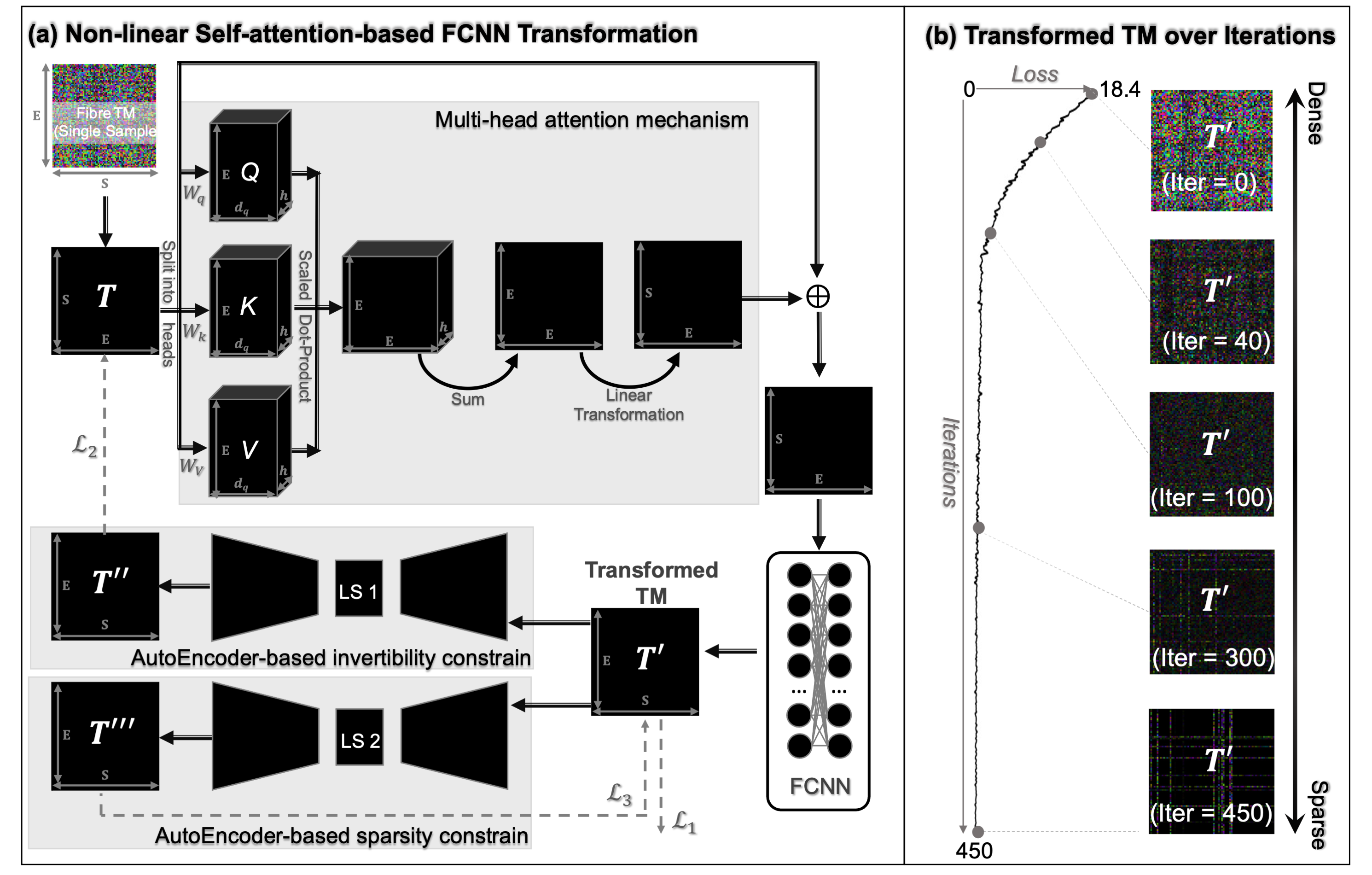}
    \caption{Self-attention-based FCNN Transformation Framework. (a) Framework overview: Multi-head attention mechanism, where Query (Q), Key (K) and Value (V) encode the input information by multiplication with trainable weight matrices $\mathbf{W_Q}$,$\mathbf{W_K}$ and $\mathbf{W_V}$. AutoEncoder-based invertibility constrain to preserve the matrix structure during training by minimising the reconstruction loss of $\mathbf{T}$ and $\mathbf{T''}$ ($\mathcal{L}_2$). AutoEncoder-based sparsity constrain to maximally compress the transformed matrices while minimising reconstruction loss with the transformed TM ($\mathcal{L}_3$). $\mathcal{L}_2$ and $\mathcal{L}_3$ are both global-phase independent loss functions (b) Example transformed TM result over 450 iterations during the model training.}
    \label{fig:attention}
\end{figure}

We first show that a dense fibre TM can be transformed into sparse representations using a self-attention-based network (Figure\ref{fig:attention} (a)). 
To demonstrate our first criterion of memory efficient representation, we apply the `Compression is Learning' philosophy \cite{deletang2023language} and maximise the sparsity of the transformed TM representation by minimising the $\ell_1$ norm. We then validate the effectiveness of the basis transformation by assessing the sparsity of the transformed representation of TMs, which can be measured by participation ratio, $p$, (see Eqn \ref{metric1}) or by the latent space size of an autoencoder (see Eqn\ref{metric2}). Generally, a greater participation ratio represents a less sparse matrix.

In language modelling, text is represented as a sequence of words where each word is a vector that forms a row of a matrix. By analogy, measured TMs can be considered a sequence of measurements performed on a fibre. In the formulation of $\mathbf{T}$ given above, each column of the fibre TM represents a different mode in the measurement basis so we use the transpose the fibre TM as the input. This input is fed into a multi-head attention mechanism by multiplying each TM by a corresponding weight vector (i.e. $\mathbf{W_Q}$, $\mathbf{W_K}$ and $\mathbf{W_V}$) to produce Query ($\mathbf{Q}$), Key ($\mathbf{K}$) and Value ($\mathbf{V}$) matrices. This process is repeated for multiple different weight matrices to achieve multihead behaviour, with the results concatenated.  This enables several different pair-wise mappings to be learned simultaneously, for example long-range and short-range correlations. The multi-head attention mechanism is residually connected to the input, $\mathbf{T}$, followed by two fully connected layers, which produce the transformed TM.

An example of a transformed TM predicted by the self-attention-based FCNN model at different iterations is shown inset in Figure \ref{fig:attention}(b). It can be seen that the transformed TM becomes rapidly less dense from 0 to 100 iterations and approaches convergence by 450 iterations.

\subsection{Validation of the sparsity of transformed TMs}
We next compare the sparsity achieved by different network architectures across multiple transmission matrix datasets to demonstrate compact representations across datasets and measurement bases. To do this, we test on four separate data sets that represent realistic TM recovery scenarios and TM scenarios. First is a set of randomly-generated forward TMs, $\mathbf{T_f}\in\mathbb{C}^{78\times 78}$, that are diagonal with one or two subdiagonals and thus fairly sparse. Second, is randomly-generated round-trip TMs representing the previous case but with a random reflector on the fibre tip to enable single-ended TM recovery \cite{zheng2023single,gordon2019characterizing}, which are dense matrices.  Third, is a set of TMs generated using a full physical model of a bent fibre of different lengths, $\mathbf{T_r}\in\mathbb{C}^{78\times 78}$, also presenting a sparse case. These first three cases are measured in an LP mode basis.  Finally, we use experimentally measured TMs recorded under a representative range of likely fibre conformations \cite{wen2023single,zheng2023single}, measured in a Fourier basis. (Details of datasets are included in Section Methods 2.1).

\begin{figure}[!htbp]
    \centering
    \includegraphics[width=1\linewidth]{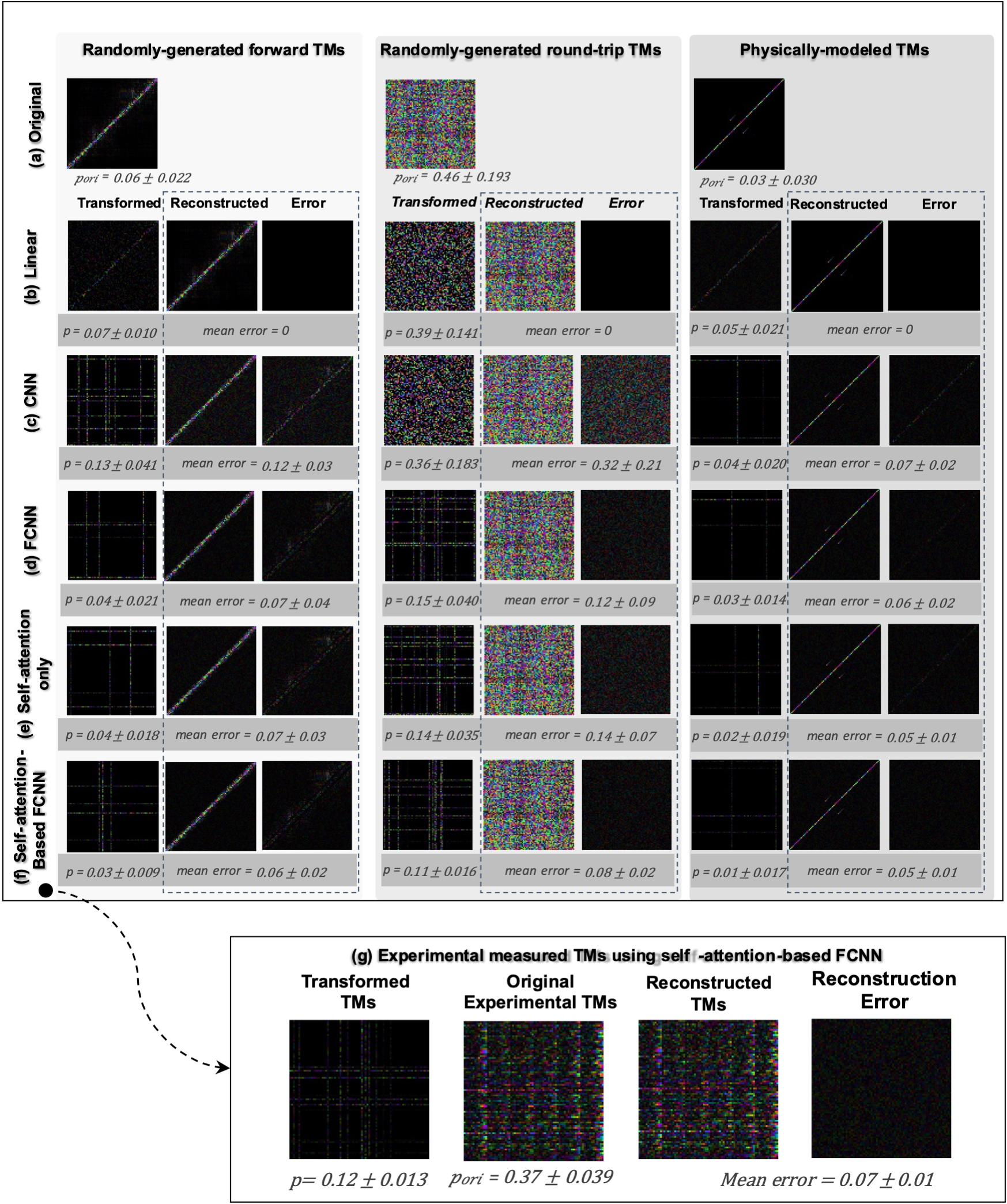}
    \caption{Transformed TMs from various datasets: Randomly-generated forward TMs, Randomly-generated round-trip TMs, physically-modeled TMs and experimental measured TMs from (a) original basis and reconstructed TMs from transformed TMs using (b) Linear similarity transformation based model, (c) CNN, (d) FCNN, (e) self-attention only, (f) self-attention-based FCNN model.(g) shows experimental measured TMs transformed from the original basis and reconstructed TMs from transformed TMs using a self-attention-based FCNN model. This model, previously validated across three datasets, demonstrates both high efficiency and accuracy, making it a robust tool for validating the experimental TMs. Metrics are quoted as mean $\pm$ standard deviation.}
    \label{fig:result12}
\end{figure}

To demonstrate that our model provides advantages over other common architectures we compare performance of these 4 datasets across five different models: averaged linear similarity transform, CNN, FCNN, self-attention only, and self-attention + FCNN (details of models given in Section Methods 2.2). The results are shown in Figure \ref{fig:result12}.

The linear model shows limited capability in transforming matrices into a more sparse basis compared to the non-linear models, exhibiting mean $p$ values similar to the original matrices (0.07 vs. 0.06 for sparse matrices and 0.39 vs. 0.46 for dense matrices). Compared to FCNN, CNN presents a significantly larger mean $p$ value, especially for dense matrices where the $p$ value is two times larger than that of using FCNN. This demonstrates that the CNN, which encodes a strong inductive bias assuming correlation between neighbouring elements, has limited capability for handling non-local correlations such as our dense fibre TMs. The self-attention-only model, which learns position-wise weights, provides improved performance over FCNN, with a further reduced mean $p$. This is because the attention mechanism selectively focuses on relevant parts of the matrices by computing alignment scores, allowing the model to dynamically prioritize information based on the input itself, whereas fully connected layers uniformly apply the same transformation to the entire input data and tend to lack the ability to shift focus for different inputs.

Finally, the self-attention + FCNN model integrates the attention mechanism before the FCNN framework and further decreases the mean $p$ value to 0.03 for less sparse matrices and 0.11 for dense matrices. 


\subsection{Validation of the invertibility of transformed TMs}
We then evaluate whether the sparse representations preserve sufficient information to reconstruct the original transmission matrices. Figure \ref{fig:result12} illustrates four examples of TMs reconstructed from the transformed TMs. The linear model provides a lossless invertible process, with a basis matrix identified by the pre-trained model. Among the non-linear models, the FCNN, the self-attention only, and the self-attention + FCNN model are all capable of recovering the TMs to the original basis with an error of $\leq 8\%$ for sparse matrices and $\leq 14\%$ for dense matrices. The CNN presents fair reconstruction for sparse matrices, with a slightly larger error of 12\%, but exhibits a large mean error of 32\% for dense matrices, indicating a limited ability for reconstruction.

\subsection{Model compatibility for different fibre mode bases}
\begin{figure}
    \centering
    \includegraphics[width=0.95\linewidth]{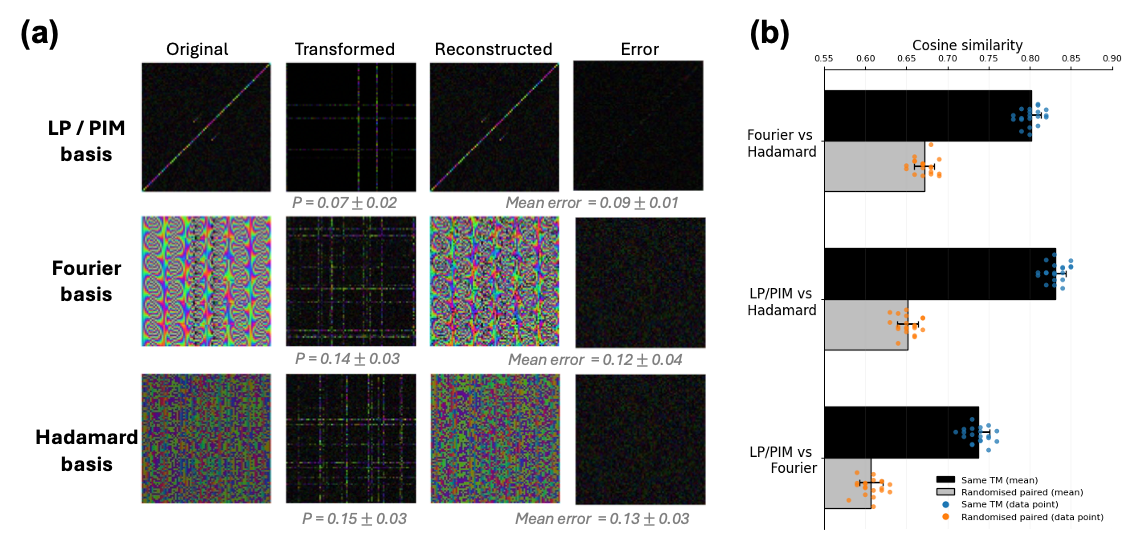}
    \caption{Basis consistency and cross-basis similarity of transformed transmission matrices. (a) Representative examples of transmission matrices expressed in three different measurement bases (LP/PIM, Fourier, and Hadamard), shown in the original domain, transformed domain, reconstructed domain, and corresponding reconstruction error. Values below the transformed matrices report the mean participation ratio $p$ (± SD), while values below the error matrices indicate the mean reconstruction error (± SD). (b) Cosine similarity between transformed transmission matrices expressed in different bases. Horizontal bars indicate the mean similarity across trials, with error bars showing standard deviation. Dots represent individual data points. Similarity is consistently higher when identical transfer matrices are used (Same TM) compared to randomly paired TMs (Randomised paired), demonstrating that the learned transformation preserves structural consistency across basis representations.}
    \label{fig:different_bases}
\end{figure}

We further examine the performance of the transformation when TMs are represented in different fibre mode bases. As shown in Figure \ref{fig:different_bases}(a), transmission matrices expressed in LP/PIM, Fourier, and Hadamard bases are all mapped to sparse representations while remaining reconstructible with low error. We then quantify the similarity between transformed TMs represented in different bases.
Figure \ref{fig:different_bases} shows that cosine similarity between transformed matrices is consistently higher when the same transfer matrix is used across bases than when matrices are randomly paired, indicating that the self-attention-based transformation preserves cross-basis structural relationships rather than overfitting to a specific basis. To further assess cross-basis consistency, we compared matched same-TM similarities with random-pair controls and baseline representations. An additional comparison with the original TMs and CNN/FCNN baselines is provided in Supplementary Figure ~S2. Across the tested basis pairs, the cosine similarity between corresponding matrices in different bases is approximately \(0.65\) for the original TMs, whereas it increases to approximately \(0.80\) for the self-attention + FCNN representation. This supports the interpretation that the learned transformation improves cross-basis structural consistency, while the appreciable random-pair similarity indicates that the metric should be interpreted comparatively rather than as evidence of complete basis invariance.

\subsection{Compatibility of experimentally-measured TMs}
We next apply the trained self-attention-based FCNN model to experimentally measured TMs to assess compatibility with physically modelled systems using TMs generated from a physical fibre model. The model, trained using a set of physically modelled perturbed transmission matrices, is applied without retraining to the experimental data. Figure~\ref{fig:result12}(g) shows the transformed representations of the experimentally measured transmission matrices in the original basis. A substantial reduction in the participation ratio is observed, with the mean $p$ value decreasing from 0.37 to 0.12, indicating that the learned transformation generalises to experimentally measured transmission matrices.

\subsection{Effect of dynamic TM perturbation on basis transformation}
\begin{figure}[!htbp]
    \centering
    \includegraphics[width=0.7\linewidth]{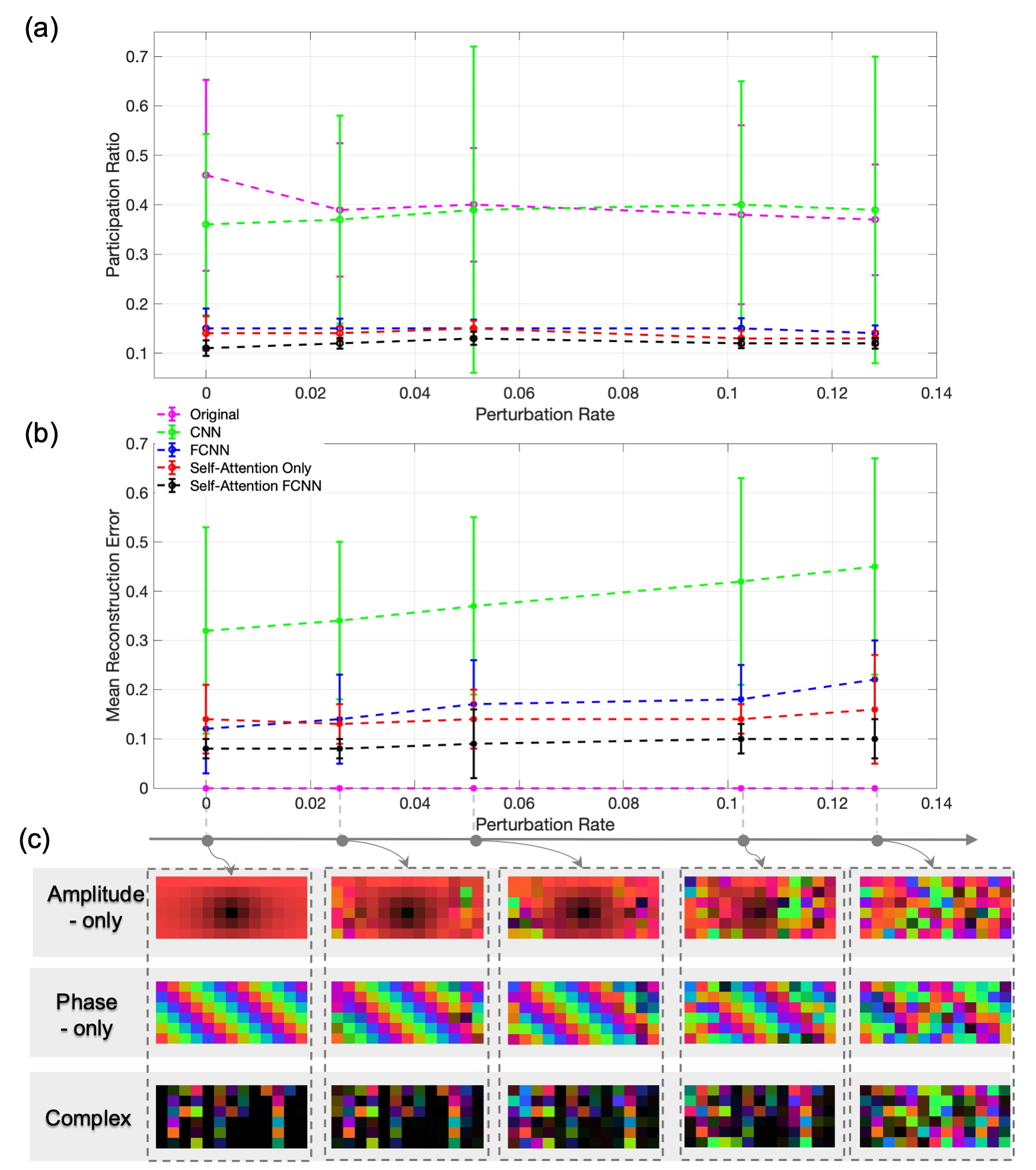}
    \caption{Effect of dynamic TM perturbation on basis transformation. (a) Participation ratio of transformed TMs from perturbed TMs with various perturbation rates using different models. (b) Mean error of reconstructed TMs from transformed TMs with various perturbation rates using different models, namely CNN (green line), FCNN (blue line), self-attention-onle (red line), self-attention-based FCNN (black line) and original matrix (purple line). (c) Three example images describing the effect of perturbation rates on fibre imaging: an amplitude-only image with a ‘radial gradient’ pattern, a phase-only diagonal gradient with a uniform amplitude and a random complex-valued image.}
    \label{fig:perturbedTM}
\end{figure}

Next, we assess the robustness of the transformation under dynamic perturbations of the transmission matrix. A situation in which the fibre is perturbed midway through TM characterisation is simulated to examine whether our model learns any error correction capability. To do this, we consider the most challenging case, our dense round-trip matrices, that also represents a realistic imaging configuration. In terms of the perturbation, we begin with a simulated round-trip TMs but replace the final few columns with data generated using a different forward TM. Specifically, we simulated and recorded round-trip TMs with four different perturbation rates, indicating the numbers of columns swapped (2/78, 4/78, 8/78, 10/78). This perturbation rate should be interpreted as the fraction of affected sequential measurements, not as a direct physical bend magnitude. This models an acquisition-time inconsistency in which the fibre moves partway through a sequential TM measurement, so that some columns are drawn from one fibre conformation while the remaining columns are drawn from a nearby perturbed conformation. The resulting matrix is therefore a hybrid of two nearby TMs, rather than the TM of a single static fibre state. Figure \ref{fig:perturbedTM} compares the robustness of different transformation models (i.e. CNN, FCNN, Self-attention only and self-attention-based FCNN) that are introduced in Section 4.2 to transform and reconstruct round-trip TMs at each perturbation rate. It can be seen that  CNN model exhibits large p values for transformed TMs and large mean error for reconstructed error, which shows low capability to achieve non-linear transformation. FCNN and the Self-attention-only model both show relatively low p values under 0.15 and mean error under 22\% with perturbation rates under 13\%. For self-attention-based FCNN model, it shows the representation of TMs transformed from the original basis (i.e. original perturbed TMs), with the lowest mean p values below 0.13 in all perturbation rates.

Realistic imaging configuration can be proposed regarding different perturbation rates as shown in Figure \ref{fig:perturbedTM}(c), where three example images with size of $6\times13$ can be recovered from perturbed fibre at each perturbation rate: an amplitude-only image with a ‘radial gradient’ pattern, a phase-only diagonal gradient with a uniform amplitude and a random complex-valued image. Random noise is added to better simulate expected behaviour in a real system. It can be seen that images can't be reconstructed to an ideal performance when the perturbation rate is larger than 10\%. This means in a practical scenario, TMs with a small perturbation rate (below 10\%) can be effectively transformed into the basis which is sparse and thus be used for fibre imaging.

\subsection{Ablation analysis}
Finally, an ablation analysis was conducted to evaluate the influence of key architectural and training hyperparameters, as summarised in Table~\ref{table1}. The results show that the autoencoder-based sparsity loss ($\mathcal{L}_2$) is necessary to maintain sparse representations in the transformed transmission matrices. Reducing the number of dense layers or the number of attention heads leads to underfitting, while increasing the network depth results in overfitting. Among the tested configurations, model~\#8 achieves the lowest participation ratio and LS ratio, indicating an improved balance between sparsity and generalisation.

To assess whether the performance gain is attributable to the attention mechanism rather than simply to increased model capacity, we include a model-complexity and runtime comparison in Supplementary Table S1. This comparison shows that increasing FCNN depth alone does not reproduce the sparsity and reconstruction trade-off achieved by the self-attention + FCNN architecture, even when the deeper FCNN contains more trainable parameters. This indicates that the attention block contributes through an input-dependent transformation rather than sorely increasing model capacity.

\begin{table}[!htbp]
\label{table1}
\caption{{Ablation analysis of hyperparameters}}
\resizebox{\textwidth}{!}{%
\begin{tabular}{|c|cc|c|cc|c|}
\hline
 &
  \multicolumn{2}{c|}{{\color[HTML]{680100} \textbf{Non-linear transformation model architecture}}} &
  {\color[HTML]{680100} \textbf{Loss function}} &
  \multicolumn{2}{c|}{{\color[HTML]{680100} \textbf{Results}}} &
  {\color[HTML]{680100} } \\ \cline{2-6}
\multirow{-2}{*}{} &
  \multicolumn{1}{c|}{\textbf{Attention layer}} &
  \textbf{FCNN} &
  \textbf{Sparsity-constrain loss} &
  \multicolumn{1}{c|}{{\color[HTML]{002060} \textbf{p}}} &
  {\color[HTML]{002060} \textbf{LS ratio}} &
  \multirow{-2}{*}{{\color[HTML]{680100} \textbf{Reconstruction}}} \\ \hline
\textit{\textbf{1}} &
  \multicolumn{1}{c|}{$\times$} &
  2 Dense layers &
  $\mathcal{L}_1$ &
  \multicolumn{1}{c|}{{\color[HTML]{002060} 0.00001}} &
  {\color[HTML]{002060} 66.50\%} &
  0.32 ± 0.21 \\ \hline
\rowcolor[HTML]{ECF4FF} 
\textit{\textbf{2}} &
  \multicolumn{1}{c|}{\cellcolor[HTML]{ECF4FF}$\times$} &
  2 Dense layers &
  $\mathcal{L}_1+\mathcal{L}_2$ &
  \multicolumn{1}{c|}{\cellcolor[HTML]{ECF4FF}{\color[HTML]{002060} 0.15 ± 0.040}} &
  {\color[HTML]{002060} 16.83\%} &
  {\color[HTML]{002060} 0.12 ± 0.09 } \\ \hline
\textit{\textbf{3}} &
  \multicolumn{1}{c|}{$\times$} &
  1 Dense layers &
  $\mathcal{L}_1+\mathcal{L}_2$ &
  \multicolumn{1}{c|}{{\color[HTML]{002060} 0.18 ± 0.022}} &
  {\color[HTML]{002060} 28.76\%} &
  {\color[HTML]{002060} 0.16 ± 0.08 } \\ \hline
\textit{\textbf{4}} &
  \multicolumn{1}{c|}{$\times$} &
  8 Dense layers &
  $\mathcal{L}_1+\mathcal{L}_2$ &
  \multicolumn{1}{c|}{{\color[HTML]{002060} 0.19 ± 0.036}} &
  {\color[HTML]{002060} 33.27\%} &
  {\color[HTML]{002060} 0.18 ± 0.10 } \\ \hline
\textit{\textbf{5}} &
  \multicolumn{1}{c|}{Multi-heads = 1} &
  $\times$ &
  $\mathcal{L}_1+\mathcal{L}_2$ &
  \multicolumn{1}{c|}{{\color[HTML]{002060} 0.21 ± 0.074}} &
  {\color[HTML]{002060} 41.09\%} &
  {\color[HTML]{002060} 0.20 ± 0.10 } \\ \hline
\rowcolor[HTML]{ECF4FF} 
\textit{\textbf{6}} &
  \multicolumn{1}{c|}{\cellcolor[HTML]{ECF4FF}Multi-heads = 2} &
  $\times$ &
  $\mathcal{L}_1+\mathcal{L}_2$ &
  \multicolumn{1}{c|}{\cellcolor[HTML]{ECF4FF}{\color[HTML]{002060} 0.14 ± 0.035}} &
  {\color[HTML]{002060} 16.83\%} &
  {\color[HTML]{002060} 0.14 ± 0.07 } \\ \hline
\rowcolor[HTML]{ECF4FF} 
\textit{\textbf{7}} &
  \multicolumn{1}{c|}{\cellcolor[HTML]{ECF4FF}Multi-heads = 2} &
  2 Dense layers &
  $\mathcal{L}_1$ &
  \multicolumn{1}{c|}{\cellcolor[HTML]{ECF4FF}{\color[HTML]{002060} 0.00001}} &
  {\color[HTML]{002060} 50\%} &
  {\color[HTML]{002060} 0.89 ± 0.10 } \\ \hline
{\cellcolor[HTML]{ECF4FF}{\color[HTML]{002060} \textit{\textbf{8}}}} &
  \multicolumn{1}{c|}{\cellcolor[HTML]{ECF4FF}Multi-heads = 2} &
  {\cellcolor[HTML]{ECF4FF}{\color[HTML]{002060} 2 Dense layers}}&
  {\cellcolor[HTML]{ECF4FF}{\color[HTML]{002060} $\mathcal{L}_1+\mathcal{L}_2$}} &
  \multicolumn{1}{c|}{\cellcolor[HTML]{ECF4FF}{\color[HTML]{002060} 0.11 ± 0.010}} &
  {\cellcolor[HTML]{ECF4FF}{\color[HTML]{002060}  14.79\%}} &
  {\cellcolor[HTML]{ECF4FF}{\color[HTML]{002060} 0.08 ± 0.02 }} \\ \hline
\end{tabular}%
}
\end{table}
\section{DISCUSSION}
We present a novel method for learning the essential features of dynamic fibre transmission matrices (TMs), yielding compressed latent-space representations that are both maximally sparse and invariant to measurement basis. Our approach is evaluated across a diverse range of TM datasets, including synthetically generated, physically simulated, and experimentally measured matrices—all under dynamic conditions and varying levels of sparsity and basis representations. Table \ref{table} summarizes the overall performance of five proposed models, validated on three distinct datasets. Metrics include the compressibility ratio (quantified by the participation ratio, pp, defined in Eqn \ref{metric1}), the latent space dimensionality from an autoencoder (Eqn \ref{metric2}), and invertibility, assessed via reconstruction error (Eqn \ref{metric3}).

It is evident that nonlinear models consistently outperform the linear baseline. As expected, participation ratio values $p$ scale with matrix dimensionality, with the theoretical lower bound given by a perfectly diagonal TM, where $p = 1/N$ for matrix dimension $N$. Among all models, the self-attention + FCNN architecture demonstrates the best performance, achieving both lower reconstruction errors and minimal participation ratios. Specifically, it reaches $p = 0.01 \pm 0.017$, approaching the theoretical limit of $p = 0.013$ for a diagonal TM. This highlights the model's ability to learn near-optimal latent representations across a wide range of TM structures and complexities. Furthermore, our approach meets the criteria established in the Introduction: it enables memory-efficient compression into sparse representations, supports complex-valued matrices, generalises across arbitrary measurement bases, and offers sufficient expressivity to model dynamic perturbations.

\begin{table}[!htbp]
\label{table}
\caption{Comparison of transformation performance across diverse datasets using different models}
\resizebox{\textwidth}{!}{%
\begin{tabular}{|c|c|l|c|c|c|c|c|}
\hline
 &
  \textit{Data Properties} &
  \multicolumn{1}{c|}{{\color[HTML]{9B9B9B} \textit{Metric}}} &
  \textbf{Linear} &
  \textbf{CNN} &
  \textbf{FCNN} &
  \textbf{\begin{tabular}[c]{@{}c@{}} Self-Attention\\ only \end{tabular}} &
  \textbf{\begin{tabular}[c]{@{}c@{}}Self-attention\\ -based FCNN\end{tabular}} \\ \hline
 &
   &
  \cellcolor[HTML]{FFFFFF}{\color[HTML]{9B9B9B} $p$} &
  \cellcolor[HTML]{FFFFFF}0.07$\pm$0.010 &
  \cellcolor[HTML]{FFFFFF}0.13$\pm$0.041 &
  \cellcolor[HTML]{FFFFFF}0.04$\pm$0.021 &
  \cellcolor[HTML]{FFFFFF}0.04$\pm$0.018 &
  \cellcolor[HTML]{ECF4FF}0.03$\pm$0.009 \\ \cline{3-8} 
 &
   &
  \cellcolor[HTML]{FFFFFF}{\color[HTML]{9B9B9B} LS ratio} &
  \cellcolor[HTML]{FFFFFF}NA &
  \cellcolor[HTML]{FFFFFF}14.79\% &
  \cellcolor[HTML]{FFFFFF}5.33\% &
  \cellcolor[HTML]{FFFFFF}5.33\% &
  \cellcolor[HTML]{ECF4FF}4.21\% \\ \cline{3-8} 
\multirow{-3}{*}{\textbf{\begin{tabular}[c]{@{}c@{}}Dataset 1: Randomly-generated \\ forward TMs\end{tabular}}} &
  \multirow{-3}{*}{\textit{Less sparse}} &
  \cellcolor[HTML]{FFFFFF}{\color[HTML]{9B9B9B} mean error} &
  \cellcolor[HTML]{FFFFFF}0 &
  \cellcolor[HTML]{FFFFFF}0.12$\pm$0.03 &
  \cellcolor[HTML]{FFFFFF}0.07$\pm$0.04 &
  \cellcolor[HTML]{FFFFFF}0.07$\pm$0.03 &
  \cellcolor[HTML]{ECF4FF}0.06$\pm$0.02 \\ \hline
 &
   &
  \cellcolor[HTML]{FFFFFF}{\color[HTML]{9B9B9B} $p$} &
  0.39$\pm$0.141 &
  0.36$\pm$0.183 &
  0.15$\pm$0.040 &
  0.14$\pm$0.035 &
  \cellcolor[HTML]{ECF4FF}0.11$\pm$0.016 \\ \cline{3-8} 
 &
   &
  \cellcolor[HTML]{FFFFFF}{\color[HTML]{9B9B9B} LS ratio} &
  NA &
  \textgreater 67.32\% &
  16.83\% &
  16.83\% &
  \cellcolor[HTML]{ECF4FF}14.79\% \\ \cline{3-8} 
\multirow{-3}{*}{\textbf{\begin{tabular}[c]{@{}c@{}}Dataset 2: Randomly-generated \\ round-trip TMs\end{tabular}}} &
  \multirow{-3}{*}{\textit{Dense}} &
  \cellcolor[HTML]{FFFFFF}{\color[HTML]{9B9B9B} mean error} &
  0 &
  0.32$\pm$0.21 &
  0.12$\pm$0.09 &
  0.14$\pm$0.07 &
  \cellcolor[HTML]{ECF4FF}0.08$\pm$0.02 \\ \hline
 &
   &
  \cellcolor[HTML]{FFFFFF}{\color[HTML]{9B9B9B} $p$} &
  \cellcolor[HTML]{FFFFFF}0.05$\pm$0.021 &
  \cellcolor[HTML]{FFFFFF}0.04$\pm$0.020 &
  \cellcolor[HTML]{FFFFFF}0.03$\pm$0.014 &
  \cellcolor[HTML]{FFFFFF}0.02$\pm$0.019 &
  \cellcolor[HTML]{ECF4FF}0.01$\pm$0.017 \\ \cline{3-8} 
 &
   &
  \cellcolor[HTML]{FFFFFF}{\color[HTML]{9B9B9B} LS ratio} &
  \cellcolor[HTML]{FFFFFF}NA &
  \cellcolor[HTML]{FFFFFF}7.25\% &
  \cellcolor[HTML]{FFFFFF}4.21\% &
  \cellcolor[HTML]{FFFFFF}4.21\% &
  \cellcolor[HTML]{ECF4FF}3.22\% \\ \cline{3-8} 
\multirow{-3}{*}
{\textbf{\begin{tabular}[c]{@{}c@{}}Dataset 3: Physical-modeled\\ perturbed TMs\end{tabular}}} &
  \multirow{-3}{*}{\textit{Sparse}} &
  \cellcolor[HTML]{FFFFFF}{\color[HTML]{9B9B9B} mean error} &
  \cellcolor[HTML]{FFFFFF}0 &
  \cellcolor[HTML]{FFFFFF}0.07$\pm$0.02 &
  \cellcolor[HTML]{FFFFFF}0.06$\pm$0.02 &
  \cellcolor[HTML]{FFFFFF}0.05$\pm$0.01 &
  \cellcolor[HTML]{ECF4FF}0.05$\pm$0.01 \\ \hline

\end{tabular}%
}
\end{table}

To further directly visualise compression efficiency, we present a sparsity–reconstruction trade-off analysis in Supplementary Figure. S1. Reconstruction error is plotted against participation ratio p, where lower p indicates a sparser transformed representation and lower reconstruction error indicates better invertibility. The self-attention + FCNN model achieves the lowest reconstruction error among the compared architectures while maintaining the lowest participation ratio, placing it closest to the favourable lower-left region of the plot. We also show selected LS2 bottleneck-size settings for the self-attention + FCNN model. Increasing the LS ratio reduces reconstruction error by retaining more information, whereas stronger compression increases reconstruction error. This confirms that the selected model represents a compromise between sparse latent representation and accurate reconstruction.

To understand why this architecture is effective, we draw parallels with classical linear algebra approaches. TM diagonalisation, achieving perfect sparsity, is possible by extracting the eigenvectors of a specific TM and using them as the basis. However, this represents a form of overfitting, as the solution is entirely specific to that fibre. Conversely, applying a fixed basis that is optimized for sparsity across many TMs avoids overfitting but underfits the data, since it remains static and linear. Our approach offers a third path: a data-dependent basis transformation regularized to use minimal parameters. This enables a nonlinear mapping that can efficiently model perturbations, which are expected to be highly compressible due to their limited physical degrees of freedom e.g., fibre curvature, twist, minor refractive index variations, optical defects, and temperature changes. This expectation is consistent with recent specklegram-based fibre-sensing work, where neural networks operating on multispecklegram inputs were able to predict complex fibre shape states with much of the performance benefit obtained from only a small number (3-9) of specklegram measurements \cite{cao2024fiber}.

The power of self-attention becomes clear when drawing inspiration from language models, where input word vectors are transformed into semantically rich latent spaces. Analogously, we treat transmission matrices as structured sequences, similar to sentences, where rows are analogous to word embeddings in a chosen measurement basis. Self-attention layers then learn input-dependent transformations that adapt to perturbations, unlike CNNs, which assume spatial locality and thus fail to generalize across arbitrary TM bases. In contrast, attention mechanisms are well suited to capture long-range, global correlations that characterize the structure of TMs.

We find that using multiple attention heads leads to improved performance. This may be because different heads can independently capture correlations at different spatial scales, potentially corresponding, in the context of fibre TMs, to intra- and inter-mode group coupling when a mode basis is used. Multi-head attention may therefore allow the model to more effectively align with the underlying physical structure and behaviour of optical fibres.

The transformed TMs exhibit structured rather than random sparsity. Averaged support maps \(H=\mathbb{E}_i[|T'_i|]\) across the four TM datasets (Supplementary Figure. S3) retain row/column-like crisscross patterns after averaging, indicating stable support across samples.This structure appears in the nonlinear transformation models but not in the linear similarity-transformation baseline, suggesting that it arises from the nonlinear models' learned sparse coordinate representation under the combined \(L_1\) sparsity and reconstruction constraints rather than from a conventional linear basis transform alone. Dataset-dependent differences in support intensity further indicate that the learned support is influenced by the statistics of the underlying TM ensemble. We therefore interpret the observed structure cautiously as a learned sparse coordinate system, rather than assigning direct physical meaning to individual active rows or columns. Additionally, future work will quantify the latent-space resolution limit using calibrated bend-angle or curvature increments and direct comparison with basis-induced latent variation.



Several avenues exist to further refine our approach. First, our method requires substantial training data, which can be challenging to obtain from experimentally measured TMs. This limitation can be mitigated by augmenting experimental data with simulated fibre datasets. Generative models for data augmentation and domain adaptation \cite{wirkert2017physiological,osman2022training} offer promising strategies to enable convergence using smaller experimental datasets or to synthesize additional training examples. Second, the training process remains memory-intensive due to the high dimensionality of TMs in imaging applications. However, large language models (LLMs) routinely operate with embedding dimensions exceeding 10,000 and sequence lengths beyond 5,000 tokens, suggesting that scaling our self-attention architecture to handle complex-valued matrices of size $\geq$ 3000 $\times$ 3000 is feasible. This scalability is particularly relevant for high-resolution MMF imaging, where future systems may involve very large transmission matrices with thousands to tens of thousands of guided modes. Although the present work uses smaller TMs as a proof of principle, Transformer-based architectures developed for large sequence and embedding dimensions may provide a route toward compact modelling of these much larger fibre systems.

While self-attention offers modest memory savings, its complexity still scales quadratically with the embedding dimension. Given that transmission matrices are of size $N^2 \times N^2$ for input fields of size $N \times N$, the overall computational cost grows as $\mathcal{O}(N^4)$, making very high-resolution processing challenging. To address this, emerging architectures such as Longformer \cite{beltagy2020longformer}, which combines local and global attention patterns, and BigBird \cite{zaheer2020big}, which uses sparse attention to avoid quadratic scaling, may enable efficient processing of much larger TMs. Additionally, the recently proposed Mamba architecture \cite{gu2023mamba}, which encodes attention through a state-space model, could further reduce memory requirements, though its impact on inference latency warrants investigation. Finally, our current training uses synthetic TMs with bandwidth-limited assumptions to avoid degeneracies from matrix logarithms. Future extensions could model broader spectral behaviour, leveraging the wavelength-dependent structure of fibre TMs via nonlinear spectral learning techniques \cite{lee2023efficient}.

Additional limitation of the present implementation is that the NN layers are not fully complex-valued. Although the paired real-valued representation retains the real and imaginary components of the TM, optimisation is performed using standard real-valued gradients rather than Wirtinger derivatives for complex-valued parameters. This design choice provides a simple and stable implementation and enables direct comparison across CNN, FCNN, and self-attention architectures, but it does not enforce complex-valued weight sharing, complex analyticity, or Wirtinger-gradient optimisation by construction. As a result, correlations between real and imaginary components are learned empirically from the data rather than imposed through a complex-valued model structure. Fully complex-valued dense, convolutional, and attention layers trained with Wirtinger-gradient-based optimisation may provide a more parameter-efficient and physically natural formulation for MMF TM modelling, and will be investigated in future work.

The successful reconstruction from a compact latent representation indicates that the learned transform captures recurring structure within the TM ensemble. Moreover, the consistent support patterns observed suggest that this structure is not arbitrary: different TM datasets exhibit distinct and repeatable support features, indicating that the model may encode physically relevant properties of the underlying fibre system. We do not yet interpret individual latent variables as direct physical coordinates such as bend radius, twist, or temperature. However, these results motivate future calibrated studies in which controlled perturbations are used to test whether LS2 coordinates, latent trajectories, support maps, or participation ratio vary systematically with physical fibre parameters.

In the future, this fibre TM representation architecture may enable hair-thin optical imaging systems to be near-real time by enabling fibre matrix estimation with minimal re-characterisation measurements tailored to minimise uncertainty in the latent space. It may also allow practical modelling of complex effects such as exploring generative models for predicting full TMs given partial information \cite{wilson2026using,li2021compressively}, wavelength dispersion and nonlinear characterisation measurements. This could in turn enable a wide range of broadband imaging modalities to be implemented through optical fibres.  Further, our models could be applied to learn more complex models as those of generalised scattering media including biological tissue, enabling faster and more accurate non-invasive methods of imaging deep inside the body, opening up a raft of new biomedical applications.

\section*{DATA AVAILABILITY}
The data presented in this study are available from the following source:
[DOI to be inserted later].

\section*{CODE AVAILABILITY}
The code for this study is available from the following source:
[DOI to be inserted later].

\section*{ACKNOWLEDGEMENT}
DBP thanks the European Research Council for financial support (PhotUntangle 804626). RJK was supported by an EPSRC doctoral training partnership grant EP/T518049/1. GSDG acknowledges support from a UKRI Future Leaders Fellowship (MR/T041951/1) and an EPSRC Ph.D. studentship.

\section*{CONFLICTS OF INTEREST}
The authors declare no conflicts of interest related to this work.
\bibliographystyle{unsrt}  
\bibliography{references}  
\end{document}